\documentclass{article}
\usepackage{spconf,amsmath,graphicx}
\usepackage{graphicx}
 \usepackage{url}

\title{Deep Convolutional Neural Network applied to Quality Assessment for Video tracking}
\name{Roger Gomez Nieto, Eugenio Tamura Morimitsu\thanks{Thanks to Colciencias and Pontificia Universidad Javeriana Colombia for funding.}}
\address{\{roger.gomez,tek\}@javerianacali.edu.co}

\begin{document}

\maketitle
\begin{abstract}
Surveillance videos often suffer from blur and exposure distortions that occur during acquisition and storage, which can adversely influence following automatic image analysis results on video-analytic tasks. The purpose of this paper is to deploy an algorithm that can automatically assess the presence of exposure distortion in videos. In this work we to design and build one architecture for deep learning applied to recognition of distortions in a video. The goal is to know if the video present exposure distortions. Such an algorithm could be used to enhance or restoration image or to create an object tracker distortion-aware.  

\end{abstract}
\section{Introduction}
\label{sec:intro}
One of the principal advantages of deep learning models is the remarkable generalization capabilities that they can acquire when they are trained on large-scale labeled datasets. Whereas models trained using proper machine learning methods are heavily dependent on the determination of, and discrimination capability of sophisticated training features, by contrast, deep learning models employ multiple levels of linear and nonlinear transformations to generate highly general data representations, very much decreasing dependence on the selection of features, which are often reduced merely to raw pixel values \cite{Krizhevsky2017}. In particular, deep convolutional neural networks (CNN) optimized for image recognition, and classification has exceedingly outperformed traditional methods. Open source frameworks such as TensorFlow \cite{Abadi2016} have hugely increased the accessibility of deep learning models, and their application to different image processing and analysis problems has much expanded.   

Nevertheless, until recently, there has been a limited effort directed towards marrying video distortions with deep networks. In principle, this could lead to a significantly improved performance in tasks as object tracking and anomaly activity detection. Deep learning engines offer a potentially powerful framework for achieving sought-after gains in performance; however, as we shall explain, progress has been limited by a lack of adequate amounts of distorted picture data, which are much harder to acquire than other kinds of labeled image data. Further, typical data-augmentation strategies such as those used for machine vision are of little use in this problem.      

In this research, we used one CNN (Convolutional Neural Network). Unlike traditional neural networks, CNN can be adapted to effectively process high-dimensional, raw image data such as RGB pixel values. Two key ideas underlie a convolutional layer: local connectivity and shared weights. Each output neuron of a convolutional layer is computed only on a locally connected subset of the input, called a local receptive field (drawing from vision science terminology). Generally, a CNN model consists of several convolutional layers followed by fully-connected layers. Some convolutional layers may be accompanied by pooling stages which reduce the sizes of the feature maps. The fully-connected layers are essentially traditional neural networks, where all the neurons in a previous layer are connected to every neuron in a current layer \cite{Kim2017}.

The results of this research will be applied to develop one object tracker distortion-aware. Video tracking is still a hard problem as many different and varying circumstances need to be reconciled in one algorithm \cite{2014a}. Hence, it is of interest to obtain information about the performance of video object trackers in distorted videos. Even though in datasets as VOT2017 \cite{Kristan2017} and previous versions, there are 60 challenging video sequences, little attention has been paid to the visual quality of their content and how the trackers are affected by authentic distortions acquired during the capture of the videos. In-capture distortions are naturally-occurring impairments such as texture distortions, artifacts due to \textbf{exposure} and lens limitations, focus, and color aberrations \cite{Ghadiyaram2018}.   

A well-structured study of the performance of the video trackers in videos affected by authentic distortions will contribute to the development of new and more powerful video tracking algorithms that work under non-ideal conditions. The data generated could be useful; for example, in fields like video surveillance where the cameras are exposed to uncontrolled environments. Hence, is essential to design a deep neural network capable of identifying when one exposure distortion is presented in a video, in automatically away. 


\section{Materials and methods}
\subsection{Hardware architecture}
To design and to train this network was needed to build a computer from the ground up. For Deep Learning, GPUs accelerate the processing of computations and are an integral part of the deep learning build. We used the GPU Titan XP, which contains 3840 NVIDIA® CUDA® cores running at 1.6 GHz and packs 12 TFLOPS of processing. The Titan XP has 12 GB of GDDR5X memory running at over 11 Gbps. Similarly, we deploy a CPU I7 8700K, with 14 nm technology, and frequency of 3.7 GHz, with six cores and 12 threads. We implement 40 GB RAM DDR4-2666 in this server. The motherboard used was a Z-370 with capacity for two GPU Titan XP. This Z370 motherboard has enough PCIe ports to support the number of GPUs that will be installed, as well as support all the other hardware components being chosen. 

For PSU (Power Supply Unit) we used an EVGA SuperNOVA 1200 P2 80+ PLATINUM, 1200W ECO Mode Fully Modular NVIDIA SLI Power Supply 220-P2-1200-X1. This because Deep learning can often require intensive periods of training, and the costs of running these instances should be minimized. The required watts for a given deep learning machine can be approximated by summing the watts of the GPU and CPUs while adding roughly 200 watts for the other components within the computer and variances in power consumption \cite{TawehBeysolow2017}. We installed the driver 390.25 of NVIDIA, Cuda 9.0, TensorFlow 1.7 and Keras 2.2.2. One complete installation guide for these libraries and frameworks can be found in \textbf{\url{https://tinyurl.com/yd6cj6wj}}. 
\subsection{Dataset}
The choice and consideration of database for training are essential for deep learning-based models since their performance depends highly on the size of the training set. In most picture quality databases, the distorted images are afflicted by only a single type of synthetically introduced distortion, such as JPEG compression, simulated sensor noise, or simulated blur \cite{Kim2017}.   

We used a distorted video surveillance dataset affected by in-capture distortions and acquired with four different surveillance cameras \cite{Nieto2018}. The videos in this dataset have an equal rate I/P frames: 10 fps. The video sequences have been degraded by using an H.264/AVC compression scheme at three different bitrates, resulting in three mirrored video sequences, that differ only in the level of compression. The three different bitrates were chosen to generate degradation all over the distortion scale (from subtle to very annoying). This dataset contains 8600 images with exposure distortion, so 8600 pristine images are selected. To the CNN, this images entry in FHD resolution (1920 x 1080). The dataset is publicly available in \textbf{ \url{https://tinyurl.com/y8zg4efw}}.
\subsection{pre-processing stages}

Before CNN training, we must do some pre-processing steps, such as 

\begin{itemize} 
	\item converts to floating-point 
	\item scale data to a range [0,1] 
	\item replace label data with one-hot encoded versions 
	\item reshapes samples to a 2D grid, one line per image 
	\item reshapes sample data to 4D tensor using channels last convention. Shaping the feature data into these 4D tensors is a necessary pre-processing step. It puts the data into the structure that is expected by the convolution layer that will sit at the start of our convnet. 
\end{itemize}   

\subsection{CNN training}
 \begin{table*}[t]
	\centering
	\caption{Results of CNN classification for exposure distortion in images, for various size training and test datasets.}
	\label{Table:CNN_Results}
	\begin{tabular}{c|c|c|c|c|c|c}
		\hline
	\textbf{Train-Test}& Epoch 1  	& Epoch 2 		& Epoch 3      	& Epoch 4		&Epoch 5& \textbf{Testing set}\\
		     \% 	& loss - acc	&loss - acc		& loss - acc 	& loss - acc	&loss - acc& loss - acc\\
		\hline
		0.1- 99.9	&0.8484-0.5294&4.7515 - 0.3529&3.2954 - 0.6471&1.7155 - 0.5294&0.5868 - 0.6471&0.8193 - 65.9330\\
		1 - 99		&2.0625-0.5200&1.4812 - 0.7143&1.7992 - 0.5429&0.6675 - 0.7886&0.3747 - 0.8857&0.3085 - 86.8178\\
		5 - 95		&3.0105-0.5463&0.6774 - 0.7337&0.2925 - 0.8629&0.2425 - 0.9029&0.2028 - 0.9246&0.1627 - 97.2466\\
		10 - 90		&2.3549-0.6543&0.2257 - 0.9246&0.1579 - 0.9611&0.1380 - 0.9571&0.1061 - 0.9783&0.0891 - 100.00\\
		20 - 80		&1.0635-0.8058&0.1367 - 0.9620&0.1016 - 0.9703&0.0992 - 0.9686&0.0430 - 0.9971&0.0359 - 99.8429\\
		50 -50		&0.3849-0.9074&0.0679 - 0.9869&0.0445 - 0.9921&0.0194 - 0.9998&0.0124 - 1.0000&0.0121 - 100.000\\
		80 - 20		&0.2660-0.9380&0.0277 - 0.9997&0.0135 - 1.0000&0.0081 - 1.0000&0.0057 - 1.0000&0.0053 - 100.000\\
		\hline	
	\end{tabular}
\end{table*}
Before entering the network, the images were reduced to 128 x 128 pixels. Additionally, the name is changed to the pictures to the following convention: the first three letters of the class (pri or exp), next to a dot, and then the consecutive number of the image (for example, pri.234). The maplotlib library was used to read the pictures. Two tensors are then created: one for the data and one for the labels.    

\begin{figure}
	\includegraphics[width=0.5\textwidth]{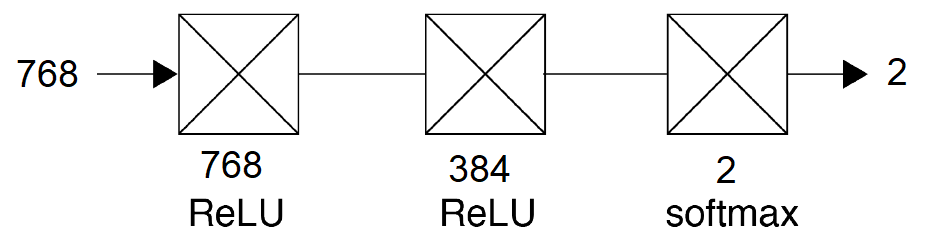}
	\caption{The architecture of our three-layer model. Each of the hidden
		layers is a dense layer with 768 neurons, one for each input. The dense layers have a ReLU activation function by default.}
	\label{Fig:CNN_Scheme}
\end{figure}

We tell Keras the overall architecture of our model is “a list of layers''.The “list of layers” architecture is called the Sequential model \cite{Glassner_2018a}. That’s perfect for us, since our architecture of Fig.~\ref{Fig:CNN_Scheme} is just three dense layers one after the other. In other words, they can be described as a 2-element list starting with the hidden layer and ending with the output layer. To build our model, we create a variable to hold a Sequential object. This sequential object is initially an empty layer of lists. Then we add our layers to that object. Each new layer takes its input from the most recently added layer. The last layer we add in is implicitly our output layer. We never explicitly say that we’re starting or ending. We add in layers until we’re done.   
We used as activation function the ReLU. ReLU is popular because it’s a fast and straightforward way to include a non-linearity step in the artificial neurons. The essential thing when using a parametric ReLU is no select a factor of precisely 1.0 because then we’d lose the kink, the function would be a straight line, and we risk that this neuron will collapse, or combine, with one that follows it.
In the last layer, we used as activation unit the softmax function. We use softmax to process the outputs of the final dense layer to turn them into probabilities. Each output neuron presents a value, or score, that corresponds to how much the network thinks that particular input belongs to the corresponding category. The larger the score, the more confident the system is that the data belongs to that category. 

As the optimizer, we used an on-line algorithm, the SGD (Stochastic Gradient Descent) because it does not require the samples to be stored or even consistent with one epoch to the next. It just handles each sample as it arrives and updates the network immediately. The algorithm SGD can take advantage of substantial efficiency gains by using the GPU for calculations, evaluating all the samples in a mini-batch in parallel \cite{Glassner_2018a}.  


There is a formula that will tell us how the quality of the match between a code and a message, by showing us the average number of bits needed to send each word in the message with that code. The number produced by that formula is the cross-entropy. The larger the cross-entropy, the more bits are required for each word. We can check how well a code would be for a given message by calculating its cross-entropy. If we compare two codes, the one that will send our message more efficiently is the one with the smaller cross-entropy \cite{Glassner_2018a}. To measure the error, we are using binary cross-entropy. That function will compare the one-hot label with the outputs from our final layer. Given that we have just two categories, and we’re using one output to decide between them (setting it to a value near 0 for one class and a value near 1 for the other), we used as the function that evaluates the error, the \textit{binary crossentropy}.  

\begin{figure}
	\includegraphics[width=0.51\textwidth]{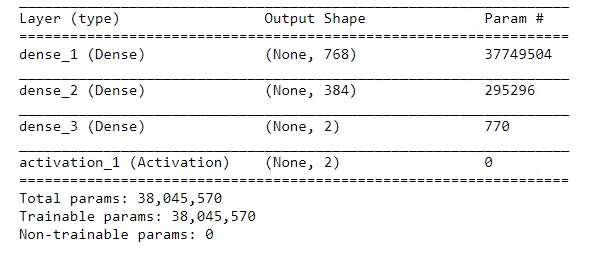}
	\caption{The architecture of our three-layer model. Each of the hidden
		layers is a dense layer with 768 neurons, one for each input. The number of parameters in this model is about 40 million, which is much lower than AlexNet and ResNet50.}
	\label{Fig:CNN_parameters}
\end{figure}

 The batch size was fixed at 128. For the direct training approach, we used the following CNN architecture: Conv-768, Conv-384 with stride 2, and FC-2. Here, “Conv” refers to convolutional layers, “FC” refers to fully-connected layers, and the trailing numbers indicate the number of feature maps (of Conv) or output nodes (of FC), as is shown in Fig. \ref{Fig:CNN_Scheme}. All of the convolutional layers were configured to use 33 filters, using zero-padding to preserve the spatial size. Following the convolutional layers, each feature map is averaged, then fed into the fully-connected layers. The number of parameters in this model is about 40 million, which is much lower than AlexNet and ResNet50, as is shown in Fig.~\ref{Fig:CNN_parameters}. This baseline model was trained using the binary cross-entropy loss in (3). The training was iterated over 5 epochs.

To evaluate the baseline models, we randomly divided each database into two subsets of non-overlapping content (distorted or pristine), with different sizes for each one.

\section{Results and discussion}

When we tried to increase the input size of the image, the GPU collapsed and returned an error. Hence, 128x128 was the ideal size for image input to CNN. The CNN trained can generate optimal results, as is detailed in Table~\ref{Table:CNN_Results}. This CNN has a high performance, even with one reduced size for train dataset. The size of the training sets is critical to the success of deep neural network models. Current public domain databases have insufficient size as compared to widely-used image recognition databases. However, constructing large-scale perceptual quality databases is a much more difficult problem than image recognition databases. The code used to deploy the CNN is publicly available in \textbf{\url{https://tinyurl.com/ydaygapw}}.

\bibliographystyle{IEEEbib}
\bibliography{ref.bib}

\end{document}